\documentclass[10pt,journal,twoside]{IEEEtran}

\IEEEoverridecommandlockouts                              

\usepackage{cite}
\usepackage[table]{xcolor}
\usepackage{soul}
\usepackage{tcolorbox} 
\soulregister\cite7
\soulregister\citep7
\soulregister\citet7
\soulregister\ref7
\soulregister\pageref7

\usepackage{amssymb}
\usepackage{bm}
\usepackage{flushend}
\usepackage{algorithm}
\usepackage{algpseudocode}

\usepackage{multirow}
\usepackage{svg}
\usepackage{float}

\usepackage[hyperfootnotes=false]{hyperref}
\usepackage{booktabs}

\hypersetup{
    colorlinks=true,
    linkcolor=black,
}

\usepackage{cuted}
\usepackage{capt-of}
\usepackage{amsmath}

\usepackage{blindtext}
\usepackage{tabularx}
\usepackage{amsmath}
\usepackage[caption=false,font=footnotesize]{subfig}
\usepackage{graphicx}

\newcolumntype{C}{>{\centering\arraybackslash}X}

\title{\LARGE \bf
OmniVLN: Omnidirectional 3D Perception and Token-Efficient LLM Reasoning for Visual-Language Navigation across Air and Ground Platforms}

\author{Zhongyuang Liu, Min He, Shaonan Yu, Xinhang Xu, Muqing Cao, Jianping Li,~\IEEEmembership{Member,~IEEE}, Jianfei Yang and Lihua Xie,~\IEEEmembership{Fellow,~IEEE}%
\thanks{This work was supported by NTUitive Gap Fund (NGF-2025-17006) and National Research Foundation, Singapore, under its Medium-Sized Center for Advanced Robotics Technology Innovation. Zhongyuang Liu is with CertaintyX. Min He, Shaonan Yu, Xinhang Xu, Jianping Li and Lihua Xie are with the School of Electrical and Electronic Engineering, Nanyang Technological University, Singapore 639798. Jianfei Yang is with the School of Mechanical and Aerospace Engineering, and jointly with the School of Electrical and Electronic Engineering at Nanyang Technological University, Singapore. Muqing Cao is with the Robotics Institute, Carnegie Mellon University, Pittsburgh, PA 15213 USA. (Zhongyuang Liu and Min He contributed equally to this work. Jianping Li is the corresponding author. E-mail: jianping.li@ntu.edu.sg)}}

\begin{document}
\maketitle
\vspace{-0.5cm}


\begin{abstract}
Language-guided embodied navigation requires an agent to interpret object-referential instructions, search across multiple rooms, localize the referenced target, and execute reliable motion toward it. Existing systems remain limited in real indoor environments because narrow field-of-view sensing exposes only a partial local scene at each step, often forcing repeated rotations, delaying target discovery, and producing fragmented spatial understanding; meanwhile, directly prompting LLMs with dense 3D maps or exhaustive object lists quickly exceeds the context budget. We present OmniVLN, a zero-shot visual-language navigation framework that couples omnidirectional 3D perception with token-efficient hierarchical reasoning for both aerial and ground robots. OmniVLN fuses a rotating LiDAR and panoramic vision into a hardware-agnostic mapping stack, incrementally constructs a five-layer Dynamic Scene Graph (DSG) from mesh geometry to room- and building-level structure, and stabilizes high-level topology through persistent-homology-based room partitioning and hybrid geometric/VLM relation verification. For navigation, the global DSG is transformed into an agent-centric 3D octant representation with multi-resolution spatial attention prompting, enabling the LLM to progressively filter candidate rooms, infer egocentric orientation, localize target objects, and emit executable navigation primitives while preserving fine local detail and compact long-range memory. Experiments show that the proposed hierarchical interface improves spatial referring accuracy from 77.27\% to 93.18\%, reduces cumulative prompt tokens by up to 61.7\% in cluttered multi-room settings, and improves navigation success by up to 11.68\% over a flat-list baseline. We will release the code and an omnidirectional multimodal dataset to support reproducible research.
\end{abstract}


\section{Introduction}

Executing object-referential instructions, such as ``find the mug near the sink,'' is a fundamental capability for embodied intelligence \cite{ahn2022saycan}. In visual-language navigation (VLN), this task requires an agent to interpret a language goal, search over a partially observed environment, disambiguate the referenced target from surrounding context, and finally convert that decision into executable motion. Recent advances in Large Language Models (LLMs) and Vision-Language Models (VLMs) have transformed navigation from a purely geometric control problem into a semantic reasoning process, where an agent must connect language, perception, memory, and action \cite{dorbala2024can, zhang2026spatialnav}. Despite this progress, reliable language-guided navigation in complex multi-room environments remains challenging, especially when long-horizon search and cross-room reasoning must be supported by real-world perception.

A first challenge lies in the sensing front-end. Existing VLN systems are commonly built upon narrow field-of-view observations, so the agent sees only a partial local scene at each step and must infer the rest from sparse glimpses \cite{chaplot2020object, yin2024sgnav}. In practice, this limited-view setting often causes repeated rotations, delayed discovery of side- or rear-located targets, and fragmented topological understanding during exploration and search. These limitations become more severe in cluttered or occluded indoor environments, where small visibility gaps can accumulate into large navigation errors across rooms. In addition, current embodied navigation systems typically rely on platform-specific sensing layouts, making it difficult to transfer a unified perception pipeline across heterogeneous aerial and ground robots. As a result, the lack of a hardware-agnostic omnidirectional perception front-end remains a key obstacle to scalable cross-platform deployment.

\begin{figure}[t]
    \centering
    \includegraphics[width=\linewidth]{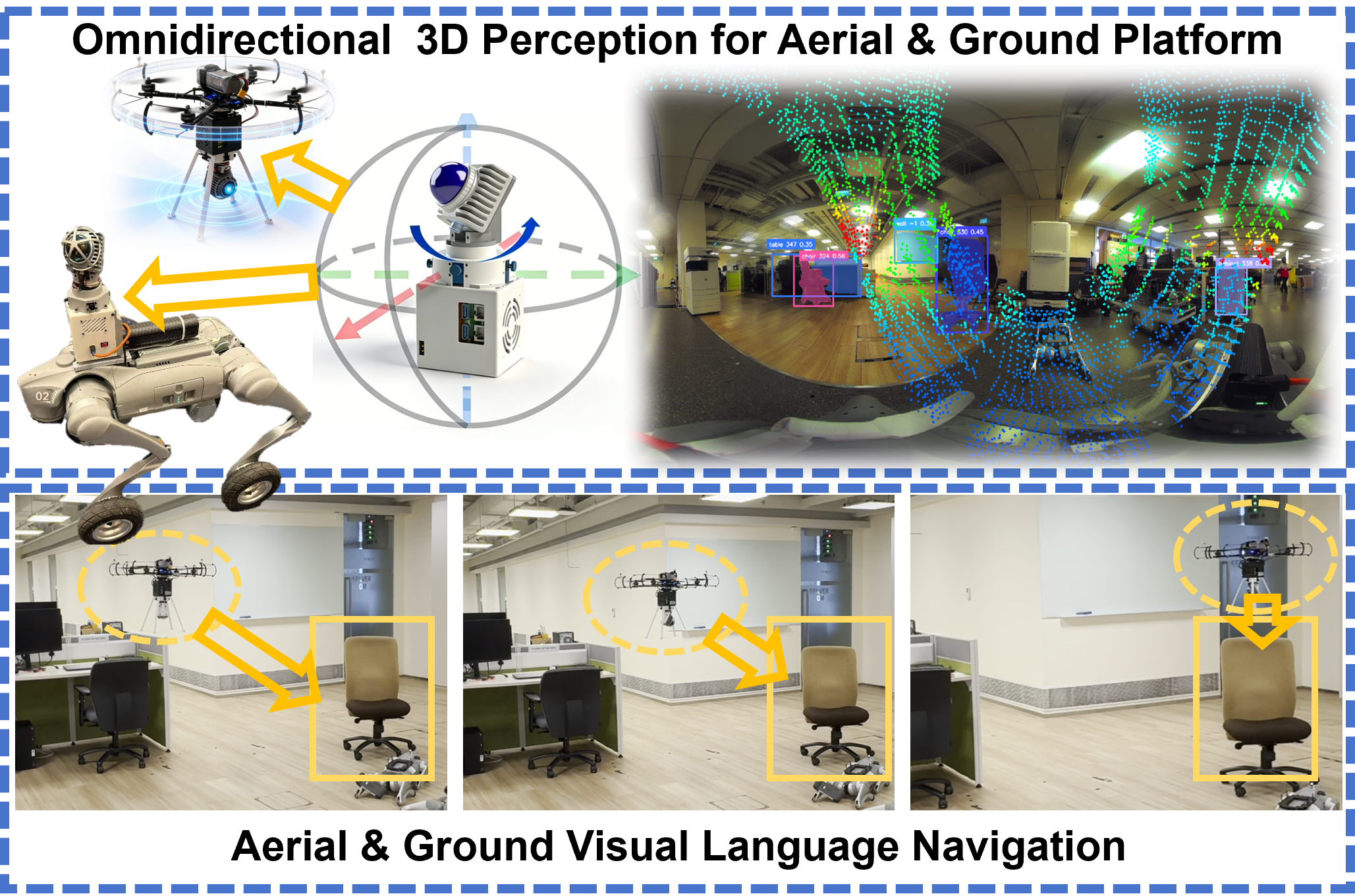}
    \caption{OmniVLN, a zero-shot visual-language navigation (VLN) framework coupling $360^\circ$ 3D perception with token-efficient reasoning across aerial \& ground platforms.}
    \label{fig:model_abstract}
\end{figure}

A second challenge arises once perception is expanded to omnidirectional 3D mapping. Although richer observations improve global situational awareness, they also produce denser semantic maps, more candidate objects, and more long-range spatial relations that must be reasoned over during navigation. Directly verbalizing such information for LLM-based planning quickly leads to a token explosion problem, where exhaustive object lists or dense semantic map descriptions exceed the prompt budget, reduce reasoning stability, and increase inference latency \cite{zantout2025sort3d, zhang2026spatialnav}. This issue is particularly harmful for long-horizon navigation, where the agent must maintain room-level context, egocentric orientation, and target-specific evidence over multiple decision steps. Therefore, an effective navigation system must not only perceive more, but also compress the world model into concise, decision-relevant representations suitable for scalable reasoning.

To address these issues, we present OmniVLN, a framework that bridges $360^\circ$ spatio-temporally consistent perception with token-efficient hierarchical reasoning. OmniVLN integrates rotating LiDAR and panoramic vision into an omnidirectional sensing and mapping stack, producing a unified semantic 3D representation that can be deployed on both aerial and ground robotic platforms. On top of this perception front-end, we incrementally construct a five-layer Dynamic Scene Graph (DSG) \cite{rosinol2021kimera}, which abstracts the environment from low-level mesh geometry to rooms and building-level structure. To support navigation-oriented reasoning, we further introduce an agent-centric 3D octant observation model that translates global graph information into an egocentric and compact spatial interface. Combined with a multi-resolution spatial attention prompting strategy, OmniVLN enables the agent to progressively select relevant rooms, orient itself with respect to the target, localize candidate objects, and emit executable navigation actions while preserving fine-grained local details and compressing distant regions into concise topological summaries. The main contribution of this paper is a unified system integration of omnidirectional 3D perception, hierarchical scene abstraction, and token-efficient LLM/VLM reasoning for language-guided navigation. Specifically, this work makes the following contributions:

(1) We develop a cross-platform omnidirectional perception front-end that combines rotating LiDAR and panoramic cameras to provide consistent 360$^\circ$ semantic 3D mapping for both aerial and ground robots.
    
(2) We build a hierarchical scene representation and reasoning pipeline that integrates Dynamic Scene Graph construction, adaptive room-level topology abstraction, and agent-centric 3D octant observations into a compact world model for cross-room embodied navigation.
    
(3) We design a practical token-efficient prompting strategy that enables scalable long-horizon reasoning over omnidirectional 3D environments without naively verbalizing dense maps or exhaustive object lists, thereby supporting target search and action generation under strict prompt budgets.
    
(4) We release an open-source multimodal dataset and deployable system framework to support reproducible evaluation of omnidirectional embodied perception and language-guided navigation.


\begin{figure*}[]
    \centering
    \includegraphics[width=0.85\textwidth]{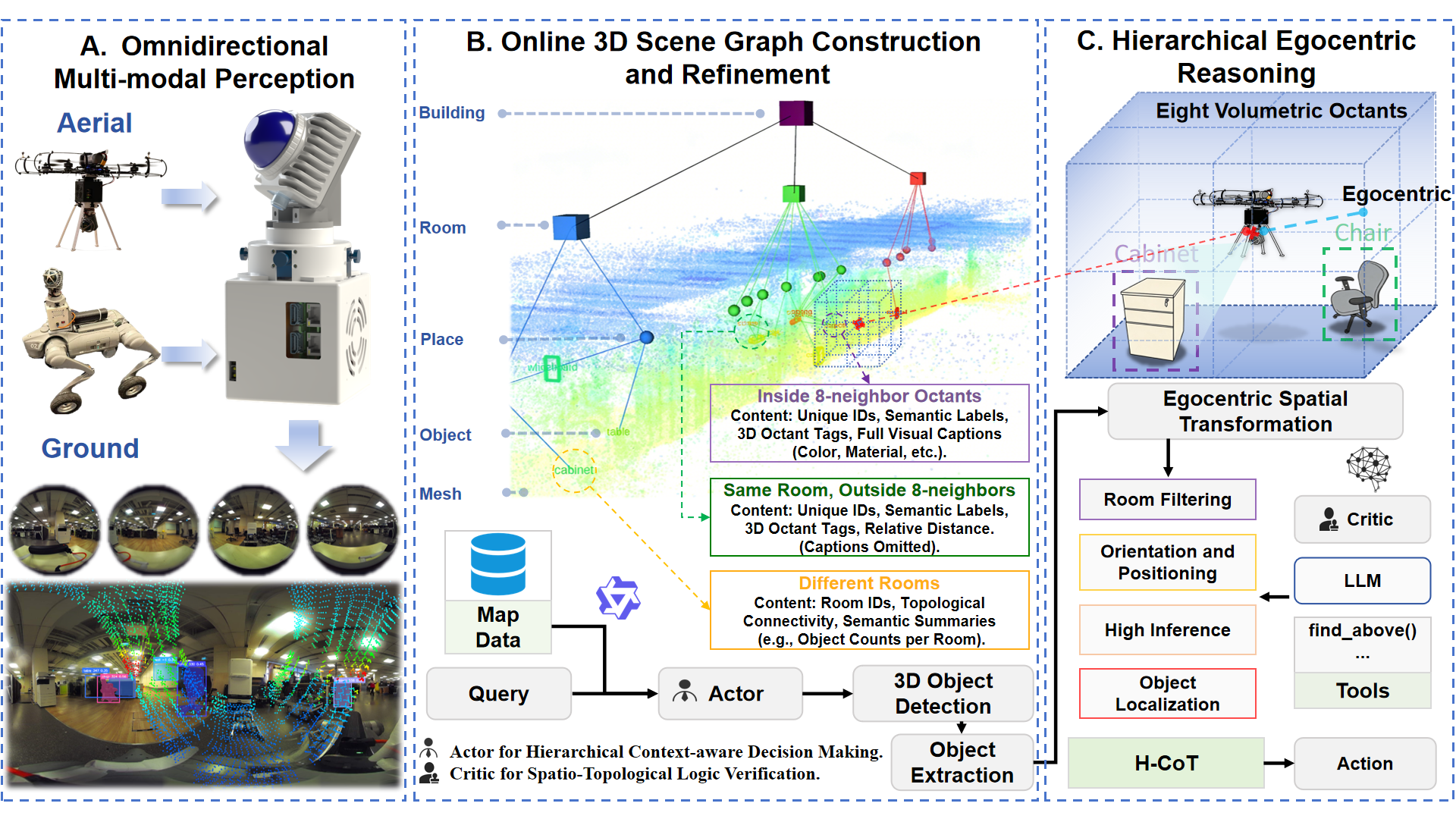}
    \caption{Overview of the proposed framework. The Multimodal Perception module (left) achieves $360^\circ$ spatio-temporal consistency by fusing data from a rotating LiDAR and panoramic fisheye cameras, enabling the generation of high-fidelity semantic point clouds across robotic platforms. The Hierarchical Representation module (center) incrementally constructs an online five-layer DSG, bridging the gap between low-level geometric places and high-level macro-spatial rooms, while employing a VLM-based hybrid pruning mechanism to refine physical veracity. The LLM-based Reasoning module (right) transforms global graph knowledge into an agent-centric 3D octant observation model, utilizing a closed-loop actor--critic framework and DSG-guided hierarchical prompting to translate natural language queries into executable navigation actions.}
    \vspace{-0.5cm}
    \label{fig:model_overview}
\end{figure*}

\section{Related Work}

\subsection{Omnidirectional Perception and Embodied Datasets}

Omnidirectional sensing improves coverage and reduces rotational overhead in cluttered indoor environments \cite{mi2021object, wang2023pano, geyer2001ijcv, scaramuzza2006iros}. Yet existing embodied systems are typically either vision-centric with limited metric fidelity or LiDAR-centric with directional blind spots \cite{Singh_2023_ICCV, geiger2012cvpr, zhang2014loam}. Mainstream navigation benchmarks also largely assume narrow-FoV RGB-D sensing \cite{savva2019habitat, deitke2020robothor, dai2017scannet, chang2017matterport3d, xia2018gibson}, leaving limited support for synchronized omnidirectional geometry, semantics, and language tasks. Our work addresses this gap with a rotating-LiDAR and panoramic-vision pipeline and a corresponding dataset.

\subsection{Hierarchical Representations and Persistent Architectures}

Hierarchical scene graphs provide a natural interface between dense 3D perception and language reasoning by compressing geometry into queryable nodes and edges \cite{hughes2023foundations, rosinol2021kimera, armeni20193d}. Prior pipelines, however, often depend on brittle room-partition heuristics and may contain physically invalid relations. Our work addresses these issues with a five-layer DSG, persistent-homology-based room partitioning, and hybrid edge validation that combines geometric pruning with VLM verification \cite{ren2024grounded, yin2024sgnav, qwen25}. The resulting persistent representation serves as a compact and physically grounded world model for planning.

\subsection{Hierarchical Egocentric Reasoning with LLMs}

LLMs are increasingly used as high-level embodied planners \cite{ahn2022saycan, zhou2023esc}, often together with VLMs for grounding and semantic verification \cite{liu2024llava, qwen25}. However, most systems still assume short-range observations and flat scene descriptions, which become unstable and expensive in large omnidirectional environments because prompt length grows with every visible object \cite{dorbala2024can, zantout2025sort3d}. Our solution couples DSG-based global abstraction with an egocentric 3D octant model and multi-resolution prompting, yielding a compact interface that preserves local detail, summarizes distant structure, and remains aligned with executable motion primitives \cite{zhang2026spatialnav, mi2025lasp, xu2025s2mllm}.


\section{OmniVLN}

This section presents the proposed OmniVLN framework. We organize the method into three components that correspond to the subsection structure: (A) Omnidirectional Multimodal Perception, which fuses a rotating LiDAR with panoramic vision to produce spatio-temporally consistent semantic 3D observations; (B) Online 3D Scene Graph Construction and Refinement, which incrementally abstracts these observations into a physically faithful five-layer DSG spanning objects, places, and rooms; and (C) Hierarchical Egocentric Reasoning with LLMs, which transforms global graph knowledge into a compact, agent-centric representation and applies multi-resolution prompting to enable scalable, tool-grounded decision making under strict context budgets.

\subsection{Omnidirectional Multimodal Perception}

\begin{figure}[h]
    \centering
    \includegraphics[width=\columnwidth, trim=30 30 30 30, clip]{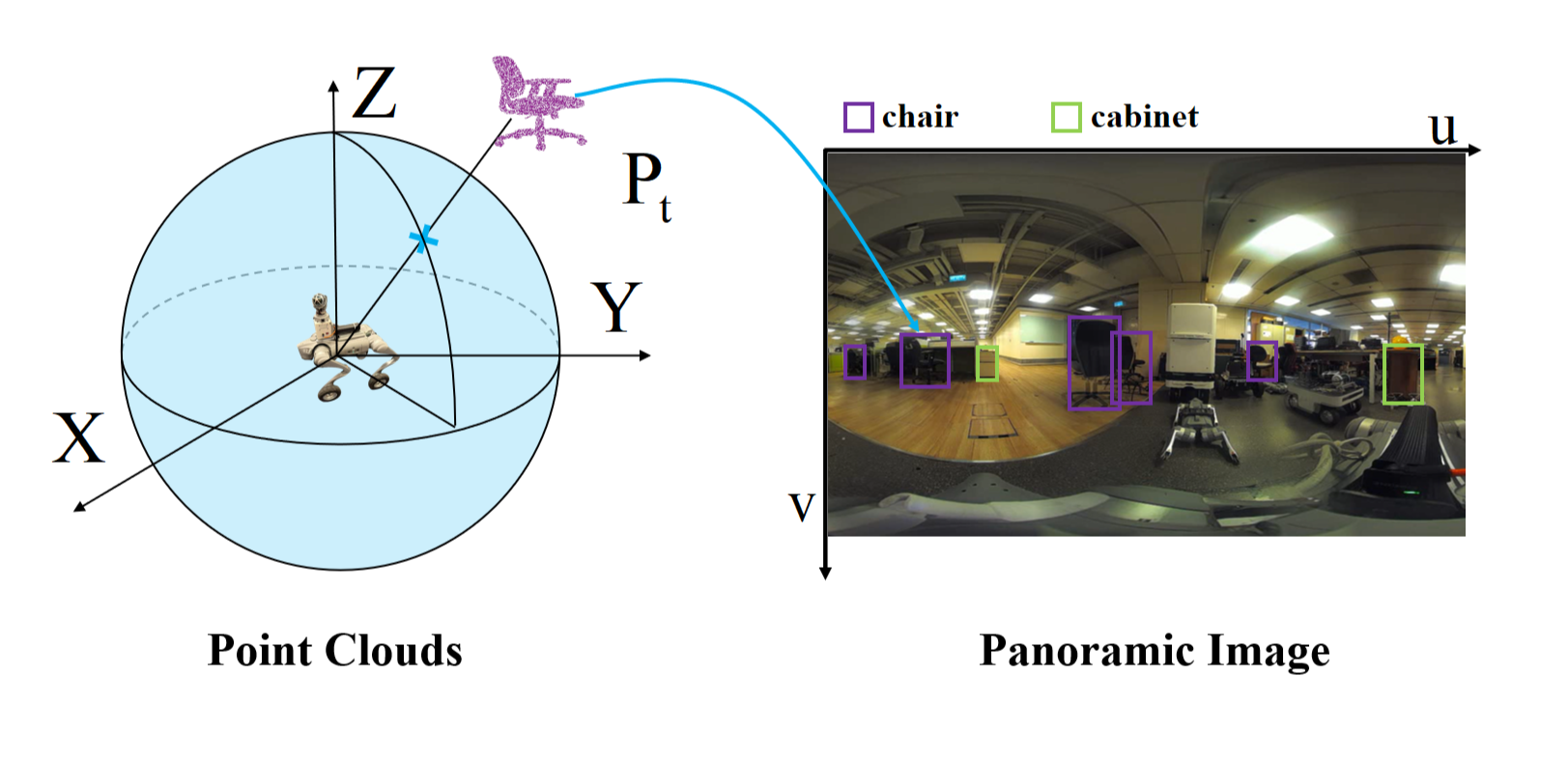}
    \caption{Omnidirectional projection model. A 3D point $P_t=(x,y,z)$ in the robot's egocentric frame is mapped onto a unit sphere and projected to pixel coordinates $(u,v)$ of a $360^\circ$ panoramic image using the equirectangular projection model.}
    \label{fig:PointCloud}
\end{figure}

Our framework achieves $360^\circ$ environmental understanding by fusing high-frequency LiDAR geometry with dense panoramic semantics. To ensure spatio-temporal consistency, raw points $P_i^{raw}$ are motion-compensated within a scan period $[t_k, t_{k+1}]$ using a LiDAR-Inertial Odometry backend \cite{xu2022fast}:
\begin{equation}
    P_i = \mathbf{T}_{w b_k} \exp \left( \hat{\omega}_k (t_i - t_k) \right) \mathbf{T}_{b c} P_i^{raw} + \mathbf{v}_k (t_i - t_k),
\end{equation}
where $\mathbf{T}_{w b_k}$, $\omega_k$, and $\mathbf{v}_k$ denote the pose and velocities at the start of the scan. This yields a globally registered scan $\mathcal{P}_{map}$.

For omnidirectional semantic awareness, we utilize a $640 \times 1920$ panoramic stream $\mathcal{I}_t$. Open-vocabulary masks $\mathcal{M}_t$ are extracted via Grounded SAM \cite{ren2024grounded} and SAM2. As shown in Fig.~\ref{fig:PointCloud}, each point $P_t = (x,y,z) \in \mathcal{P}_{map}$ is mapped to panoramic pixel coordinates $(u,v)$ through equirectangular projection:
\begin{equation}
\theta = \operatorname{atan2}(y, x), \quad \phi = \operatorname{asin}\left(z / \lVert P_t \rVert \right)
\end{equation}
\begin{equation}
u = \frac{W}{2} \left( 1 + \frac{\theta}{\pi} \right), \quad v = \frac{H}{2} \left( 1 - \frac{\phi}{\pi/2} \right)
\end{equation}
where $W$ and $H$ are image dimensions. By querying $\mathcal{M}_t$ at $(u, v)$, each point is assigned a label $L_p$. To suppress perception noise, the final label $\mathcal{L}(o_j)$ of an object node $o_j \in \mathcal{V}_{object}$ is determined by a majority vote over $T$ frames:
\begin{equation}
    \mathcal{L}(o_j) = \mathop{\arg\max}_{l \in \text{Labels}} \sum_{t=1}^T I(L_{p \in o_j, t} = l)
\end{equation}
where $I(\cdot)$ is the indicator function. This unified representation tightly couples geometry (Mesh) and instances (Objects), providing a consistent input for the hierarchical DSG construction.

\subsection{Online 3D Scene Graph Construction and Refinement}

The environment is modeled as a five-layer Hierarchical DSG $\mathcal{G} = (\mathcal{V}, \mathcal{E})$: L1 (Mesh): TSDF-based geometry; L2 (Object): SAM2-instantiated entities with centroids $\mathbf{p}_i$ and semantics $\mathcal{L}_i$; L3 (Place): GVD-based topological nodes \cite{hughes2023foundations} with clearance $R(v)$; L4 (Room): macro-spatial clusters; L5 (Building): the root node.

\subsubsection{Adaptive Room Partitioning via Persistent Homology}
To achieve threshold-free partitioning, we apply Persistent Homology (PH) \cite{yin2024sgnav} to the place graph $\mathcal{G}_P$. A filtration is defined by iteratively removing nodes with $R(v) < \delta$, yielding nested subgraphs $K_0 \subseteq K_1 \subseteq \dots \subseteq \mathcal{G}_P$. We identify the optimal room count by seeking the longest stable plateau in the 0-th Betti number ($\beta_0$) curve:
\begin{equation}
    \delta^* = \mathop{\arg\max}_{\delta_i, \delta_j} \{ | \delta_j - \delta_i | : \beta_0(\delta) = k, \forall \delta \in [\delta_i, \delta_j] \}
\end{equation}
where $k > 1$. Nodes are then clustered into $\mathcal{V}_{room}$ based on these topological components.

\subsubsection{Hybrid Edge Pruning and Refinement}
We implement a dual-stage refinement to ensure graph veracity:
1) Geometric Pruning: An edge $e_{ij}$ is pruned if the direct path is obstructed by the map $\mathcal{P}_{map}$:
\begin{equation}
    \exists \mathbf{p} \in \text{line}(\mathbf{p}_i, \mathbf{p}_j) \text{ s.t. } \text{dist}(\mathbf{p}, \mathcal{P}_{map}) < \epsilon
\end{equation}
2) VLM Verification: For short-range edges ($<1$m), we query Qwen2.5-VL \cite{qwen25} with the panoramic frame $\mathcal{I}_t$:
\begin{equation}
    e_{ij} = 
    \begin{cases} 
    1, & \text{VLM}(\mathcal{I}_t, \text{Prompt}(\mathcal{L}_i, \mathcal{L}_j)) > \tau_{vlm} \\
    0, & \text{otherwise}
    \end{cases}
\end{equation}
Functional groups are subsequently aggregated to simplify the reasoning hierarchy.

\subsubsection{Cognitive Persistence}
The framework serializes $\mathcal{G}$ and its metadata into JSON/PLY formats. This decoupling enables offline inference via a persistent digital twin, allowing zero-shot navigation and planning without active sensor streams.

\subsection{Hierarchical Egocentric Reasoning with LLMs}
We propose a reasoning framework that distills topological knowledge from the persistent DSG into actionable egocentric commands, bridging the gap between global maps and the agent’s immediate spatial context.

We formulate navigation as a closed-loop decision process that maps a language query, the current robot pose, and the persistent DSG memory to executable action primitives. At decision step $t$, the reasoning module receives the instruction $q$, the current pose $\mathbf{T}_{wb}^t$, and the current graph state $\mathcal{G}^t$, then outputs the next high-level action $a_t$ for exploration, target approach, or goal termination. The key challenge is to preserve enough room-level and object-level context for cross-room search without overwhelming the LLM with the full 3D map.

\subsubsection{3D Octant Spatial Transformation}
To provide the LLM with an intuitive first-person perspective, we transform DSG node coordinates $\mathbf{P}_i$ into an agent-centric 3D octant model. Given the pose $\mathbf{T}_{wb} = [\mathbf{R}_{wb} | \mathbf{t}_{wb}]$, the local coordinate $\mathbf{P}'_i = [x'_i, y'_i, z'_i]^\top$ is:
\begin{equation}
    \mathbf{P}'_i = \mathbf{R}_{wb}^\top (\mathbf{P}_i - \mathbf{t}_{wb})
\end{equation}
Extending the 2D spatial discretization in \cite{zhang2026spatialnav}, we partition the environment into eight volumetric octants $\mathcal{S}$—akin to a $2 \times 2$ Rubik’s cube—based on the signs of the local axes:
\begin{equation}
    \text{Oct}(o_i) = \text{sgn}(x'_i) \otimes \text{sgn}(y'_i) \otimes \text{sgn}(z'_i - h_{cam})
\end{equation}
where $h_{cam}$ is the camera height. This mapping generates discrete egocentric cues (e.g., Front-Left-Top), enabling the LLM to infer vertical alignments and immediate visibility. Importantly, the octant representation is built from omnidirectional perception and persistent DSG memory rather than a single narrow-FoV image, so nearby side and rear candidates remain available to the planner without requiring repeated exploratory rotations. We further serialize octant contents using multi-resolution summaries: nearby nodes retain instance-level detail, while peripheral and distant regions are compressed into room- and functional-group-level descriptors. This representation provides the LLM with a navigation-oriented state that is both egocentric and globally informed.

\subsubsection{DSG-guided Hierarchical Chain-of-Thought}
To transcend the 3-meter perception bottleneck typical of local-policy agents \cite{dorbala2024can}, our H-CoT mechanism guides the LLM through a top-down logical traversal of the DSG. Given the instruction $q$, the model first performs \textit{Room Filtering} to identify rooms that are topologically reachable and semantically consistent with the target description. It then performs \textit{Orientation} using the 3D octant model to determine where the next search or motion step should be directed in the agent-centric frame. Next, \textit{Functional Group Inference} exploits object co-occurrence and room context to prioritize candidate regions such as tables, shelves, or counters that are likely to contain the target. Finally, \textit{Object Localization} grounds the target to a specific node or a short list of candidate nodes for execution. This structured prompting leverages long-term DSG memory to support zero-shot, cross-room navigation while avoiding flat enumeration of all objects in the map.

Operationally, this hierarchy decomposes long-horizon navigation into a sequence of smaller decisions: where to go next, which local direction to face, and which object or waypoint to approach. Such decomposition is important in cluttered indoor scenes, where directly asking the LLM to choose from all rooms and all objects in one step is both token-inefficient and error-prone. By progressively narrowing the candidate set from room-level topology to object-level grounding, the reasoning process remains aligned with how the robot physically executes navigation.

\subsubsection{Actor-Critic and Tool-use Integration}
Decision-making is implemented via an Actor-Critic loop. The Actor generates symbolic navigation primitives such as room transition, heading adjustment, target approach, and stop commands (e.g., go\_to\_room(id), turn\_to(octant), go\_near(id), stop(id)), while the Critic verifies that the proposal is logically consistent with the current DSG, the instruction semantics, and the robot's traversability constraints. To resolve complex spatial relations, the LLM invokes specialized tools (e.g., find\_above):
\begin{equation}
    \mathcal{A}_{next} = \text{LLM}(\text{Prompt}_{H\text{-}CoT}, \mathcal{G}_{DSG}, \mathcal{S}_{oct}, \text{Tools})
\end{equation}
If the Critic detects inconsistencies, such as selecting an object outside the filtered room set or issuing an action incompatible with the current topology, the action is rejected and the Actor replans using the updated critique. The validated command is then executed by a deterministic local policy \cite{mi2025lasp}, which converts the symbolic output into platform-specific waypoints or control setpoints. After execution, the robot pose and DSG state are updated, and the next reasoning cycle begins. This closed loop allows OmniVLN to interleave global semantic search with local motion execution, yielding robust target-oriented navigation across aerial and ground platforms.


\begin{figure*}[t] 
    \centering
    \includegraphics[width=0.9\textwidth]{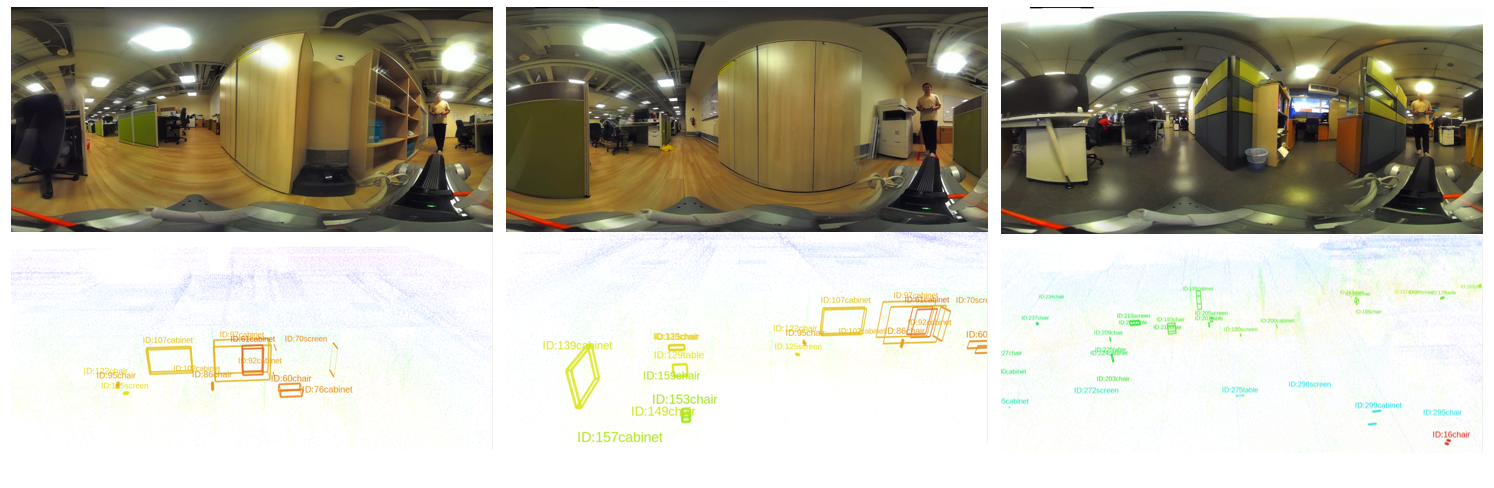}
    \caption{Visualization of the multi-modal perception in the IoT laboratory. The top row shows panoramic RGB observations from the agent's perspective. The bottom row displays the corresponding 3D semantic point clouds with instantiated objects and their unique IDs, which serve as the raw input for the DSG hierarchy construction.}
    \label{fig:iot_lab_perception}
    \vspace{-0.5cm}
\end{figure*}

\section{Experiments}

\subsection{Spatial Reasoning and Referring Expression Evaluation}

To evaluate whether the hierarchical DSG architecture improves the LLM spatial logic, We conduct a Referring Expression Generation (REG) experiment in a real-world IoT laboratory consisting of three rooms (visualized in Fig.~\ref{fig:iot_lab_perception}). The objective is to assess the model's ability to identify and describe objects based on their intrinsic properties and relative spatial contexts.

\subsubsection{Experimental Design and Prompting Strategy}

We design a dual-category evaluation metric:
\begin{itemize}
    \item View-Independent (VI) Accuracy: Measures the ability to describe an object’s inherent attributes (color, material, function) and its fixed logical position (e.g., room membership and functional group) as defined by the DSG hierarchy.
    \item View-Dependent (VD) Accuracy: Measures the reasoning capability for relative spatial relations (e.g., "to the left of," "between") by processing raw 3D coordinates $(x, y, z)$.
\end{itemize}

The LLM is prompted to act as a High-Level Spatial Robotics Expert. 
In the Hierarchical (Ours) setting, the model is provided with both 
latest\_scene\_graph.txt (DSG) and scene\_objects.json (metadata). 
It is instructed to follow a Chain-of-Thought reasoning process: first anchoring 
the queried object to its corresponding logical group (e.g., ``Meeting Table Group''), 
and then comparing its coordinates with nearby objects within that specific cluster. 

In contrast, under the Non-Hierarchical (Baseline) setting, the model receives only the flat scene\_objects.json list, requiring it to reason over all objects in the scene without any topological or structural guidance. The output is generated in a structured JSON format containing both VI and VD descriptions for each of the 44 test objects in the IoT laboratory environment. Success is defined as a description that is both logically self-consistent and physically accurate according to ground truth.

\subsubsection{Results and Cognitive Analysis}

The quantitative results of the spatial reasoning test in the IoT laboratory are summarized in Table~\ref{table:reasoning_accuracy}. The data reveals a significant performance gap between hierarchical and non-hierarchical representations across all metrics.

\begin{table}[h]
\caption{Comparative Accuracy of Referring Expression Generation}
\label{table:reasoning_accuracy}
\centering
\begin{small}
\begin{tabularx}{\columnwidth}{lCCC}
\toprule
\textbf{Representation} & \textbf{VI Accuracy} & \textbf{VD Accuracy} & \textbf{Overall} \\
\midrule
Non-Hierarchical (Uniform Flat) & 86.36\% & 68.18\% & 77.27\% \\
\textbf{Hierarchical (Ours)} & \textbf{95.45\%} & \textbf{90.91\%} & \textbf{93.18\%} \\
\midrule
\rowcolor[gray]{0.9}
\textbf{Improvement} & \textbf{+9.09\%} & \textbf{+22.73\%} & \textbf{+15.91\%} \\
\bottomrule
\end{tabularx}
\end{small}
\end{table}

The most striking observation is the 22.73\% improvement in View-Dependent (VD) reasoning. In the non-hierarchical baseline, the LLM frequently suffered from "spatial hallucination" when processing a long list of absolute 3D coordinates. Without topological boundaries, the model attempted to calculate spatial relations across the entire scene, leading to logic errors such as describing an object as being "next to" another object located in a different room. 

In contrast, our hierarchical DSG provides a cognitive heuristic that mimics human spatial partitioning. By anchoring objects within specific [Groups] and [Rooms], the DSG effectively prunes the search space. The LLM is guided to prioritize intra-group relationships, where objects are visually and functionally clustered. This "divide-and-conquer" approach reduces the complexity of coordinate comparison from a global scale to a localized, semantically relevant context, thereby nearly eliminating cross-boundary reasoning errors.

Furthermore, the 9.09\% gain in View-Independent (VI) accuracy underscores the value of explicit logical positioning. While the flat baseline must infer room membership through geometric boundaries (which are prone to errors near doorways), the DSG provides direct topological ground truth. This ensures that the robot’s descriptions remain stable and accurate regardless of the observer's viewpoint, providing a reliable semantic foundation for complex human-robot interaction.

\subsection{Token Efficiency Analysis}

To quantitatively evaluate the scalability and efficiency of our Multi-resolution 3D Octant Prompting mechanism, we conduct a series of stress tests across varying environment scales and object densities. This analysis focuses on the cumulative end-to-end token consumption, which directly impacts inference latency and deployment costs in large-scale embodied tasks.

\begin{figure}[h]
    \centering
    \includegraphics[width=\columnwidth,trim=0 0 0 0,clip]{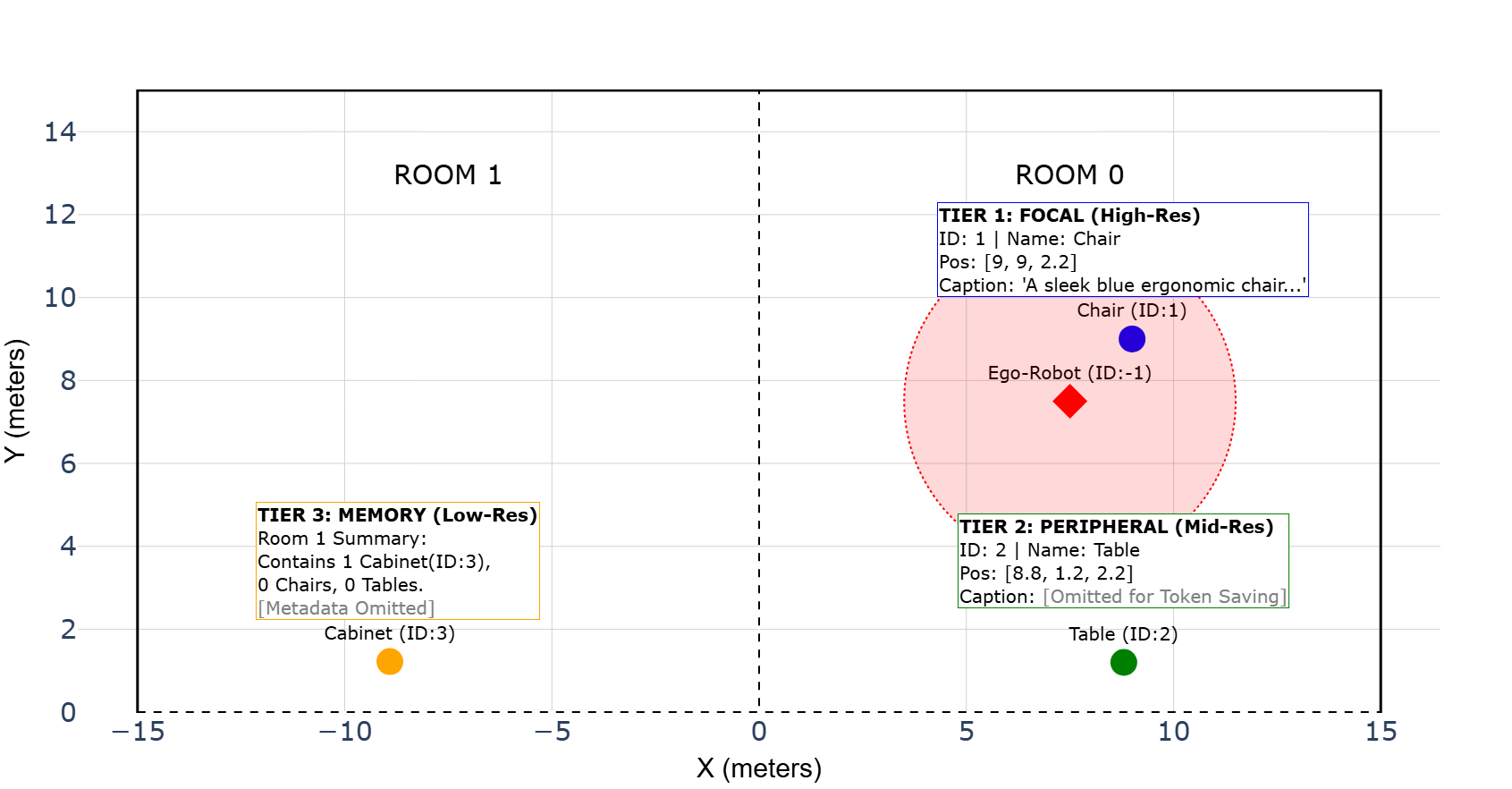}
    \caption{Multi-resolution spatial attention mechanism for token-efficient navigation. The central red diamond denotes the ego-agent. Semantic object categories are color-coded as follows: chair (blue), table (green), cabinet (orange).}
    \label{fig:token_mechanism}
\end{figure}

As illustrated in Fig.~\ref{fig:token_mechanism}, by calculating the relative Euclidean distance and room occupancy in the backend, we dynamically adjust the resolution of object descriptions. Only objects in the agent's focal zone (Tier~1) receive full semantic captions, while objects in the peripheral room (Tier~2) or distant environments (Tier~3) are progressively summarized, resulting in a dramatic reduction of input tokens without sacrificing global topological awareness.

\subsubsection{Synthetic Dataset Generation and Setup}

\begin{figure*}[t]
    \centering
    \subfloat[$D_3$: Focal tier.\label{fig:top_view_d3}]{\includegraphics[width=0.23\textwidth]{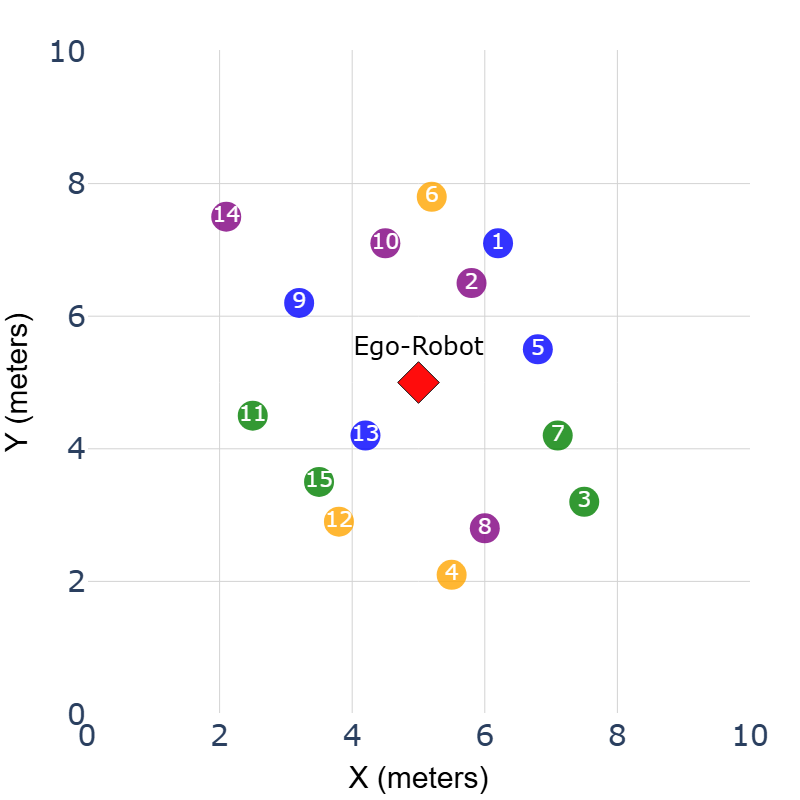}}
    \hfill
    \subfloat[$D_6$: Peripheral tier.\label{fig:top_view_d6}]{\includegraphics[width=0.23\textwidth]{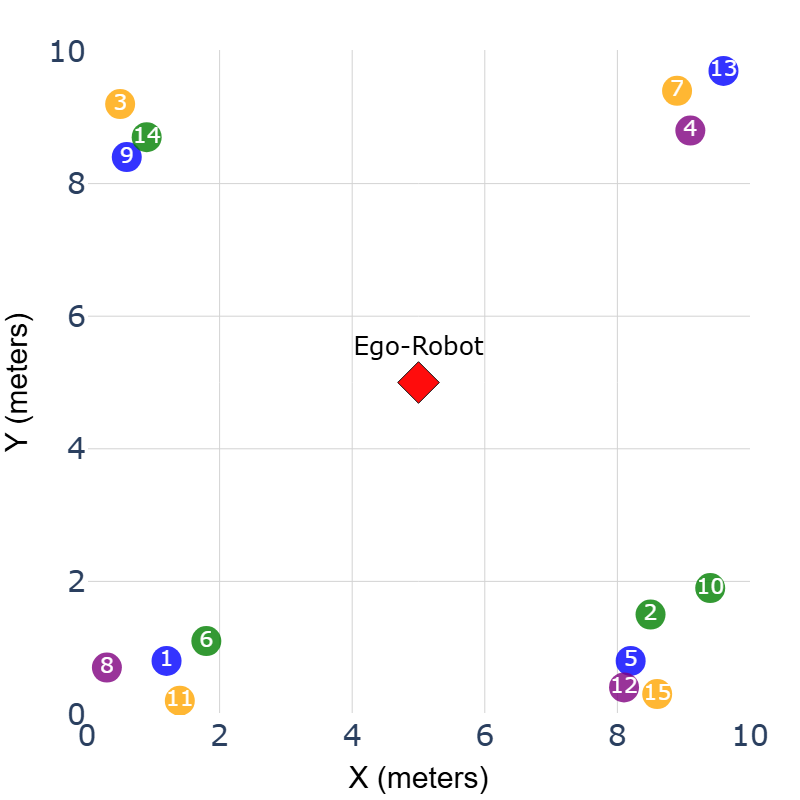}}
    \hfill
    \subfloat[$D_9$: Global-memory tier.\label{fig:top_view_d9}]{\includegraphics[width=0.23\textwidth]{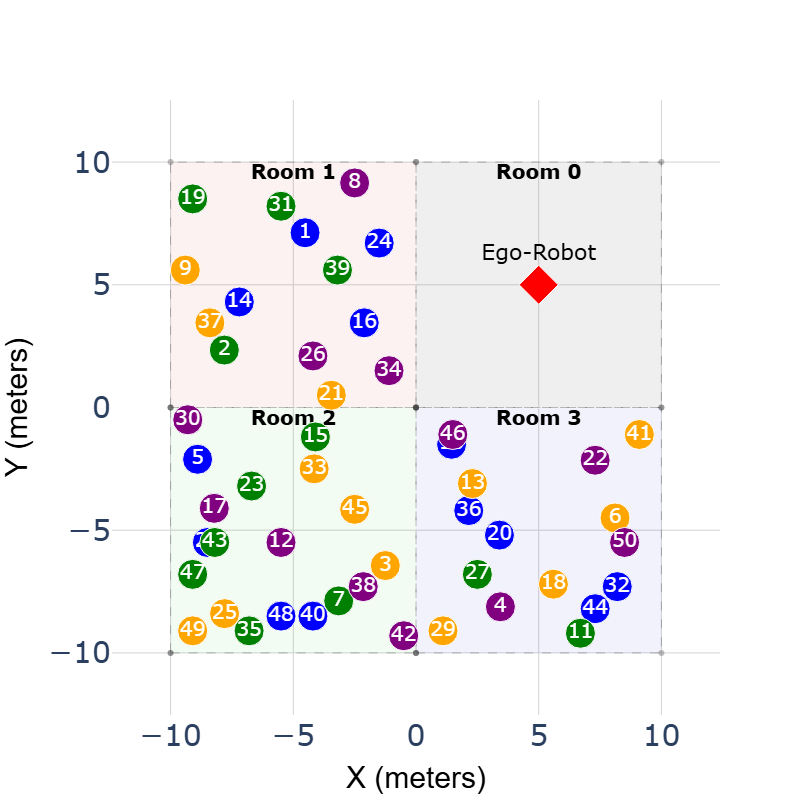}}
    \caption{Synthetic datasets across the three spatial-attention tiers. Each scene represents a $10\text{m} \times 10\text{m}$ environment with varying object densities and distributions. Colors denote chair (blue), table (green), cabinet (orange), and screen (purple).}
    \label{fig:synthetic_comparison}
\vspace{-0.25cm}
\end{figure*}

Since existing benchmarks like MP3D \cite{chang2017matterport3d} often feature fixed object densities, we developed a synthetic scene generator to produce nine distinct datasets ($\text{D}_1$ to $\text{D}_9$) with controllable spatial conditions. The simulation environment consists of four interconnected rooms, each measuring $5\text{m} \times 5\text{m}$. For consistency, the agent is initialized at the center of Room 0 at $[2.5, 2.5, 0.0]^\top$ facing forward.

Objects are randomly instantiated from four categories (chair, table, cabinet, screen) with unique IDs and detailed visual captions (20--40 words each, simulated from VLM outputs). We categorize the spatial distribution into three attention tiers based on our proposed model:

\begin{itemize}
    \item Foveal Tier ($\text{D}_1$--$\text{D}_3$): Objects are placed within the $3\text{m}$ circular focal range and distributed across all eight 3D octants.
    \item Peripheral Tier ($\text{D}_4$--$\text{D}_6$): Objects are located within the same room but outside the $3\text{m}$ focal range (e.g., $X, Y \in [0, 2]\cup[8, 10]$).
    \item Global Memory Tier ($\text{D}_7$--$\text{D}_9$): Objects are randomly distributed across distant rooms (Room 1, 2, and 3).
\end{itemize}

The density is varied from Low ($N=5, 10, 15$) to High ($N=15, 30, 50$) to simulate increasingly cluttered environments, as detailed in Fig.~\ref{fig:synthetic_comparison}. This setup allows us to isolate the impact of spatial attention on the prompt budget.

\subsubsection{Quantitative Results and Comparative Analysis}

\begin{table*}[t]
\caption{End-to-End Cumulative Token Consumption Across Different Spatial Conditions and Object Densities.}
\label{table:token_performance}
\centering
\begin{small} 
\begin{tabularx}{\textwidth}{l|CCC|CCC|CCC} 
\toprule
\textbf{Spatial Condition} & \multicolumn{3}{c|}{Foveal Tier (8-Octant)} & \multicolumn{3}{c|}{Peripheral Tier} & \multicolumn{3}{c}{Global Memory Tier} \\
\midrule
\textbf{Object Density} & Low (5) & Mid (10) & High (15) & Low (5) & Mid (10) & High (15) & Low (15) & Mid (30) & High (50) \\
\textbf{Dataset ID} & $D_1$ & $D_2$ & $D_3$ & $D_4$ & $D_5$ & $D_6$ & $D_7$ & $D_8$ & $D_9$ \\
\midrule
\textbf{Uniform Flat-List} & 402.60 & 746.80 & 1153.07 & 398.80 & 806.50 & 1181.60 & 1126.33 & 2089.33 & 3555.98 \\
\textbf{Ours} & 401.60 & 440.40 & 604.27 & 291.80 & 446.50 & 584.93 & 496.27 & 735.27 & 1067.62 \\
\midrule
\rowcolor[gray]{0.9} 
\textbf{Reduction (\%)} & \textbf{0.25\%} & \textbf{41.03\%} & \textbf{47.59\%} & \textbf{26.83\%} & \textbf{44.64\%} & \textbf{50.50\%} & \textbf{55.94\%} & \textbf{64.81\%} & \textbf{69.98\%} \\
\bottomrule
\end{tabularx}
\end{small}
\end{table*}

Cumulative token consumption is evaluated using Qwen2.5-VL-Instruct against a flat-list baseline representing raw object enumeration. Table II summarizes these results across three spatial tiers.

Our method reduces token usage across all conditions, with efficiency gains increasing alongside object density. In the foveal tier, a reduction of up to 47.59\% is observed at high densities. For the peripheral tier, our spatial attention mechanism achieves a 50.50\% reduction at D6 by filtering distant visual captions. The highest efficiency is reached in the global memory tier. At a density of 50 objects, the baseline requires 3555.98 tokens, whereas our DSG-based summary uses only 1067.62 tokens. This 69.98\% reduction facilitates building-scale scalability by preventing context overflow while preserving essential topological awareness.

\subsubsection{Navigation Performance and Robustness}

\begin{figure}[h]
    \centering
    \includegraphics[width=\columnwidth]{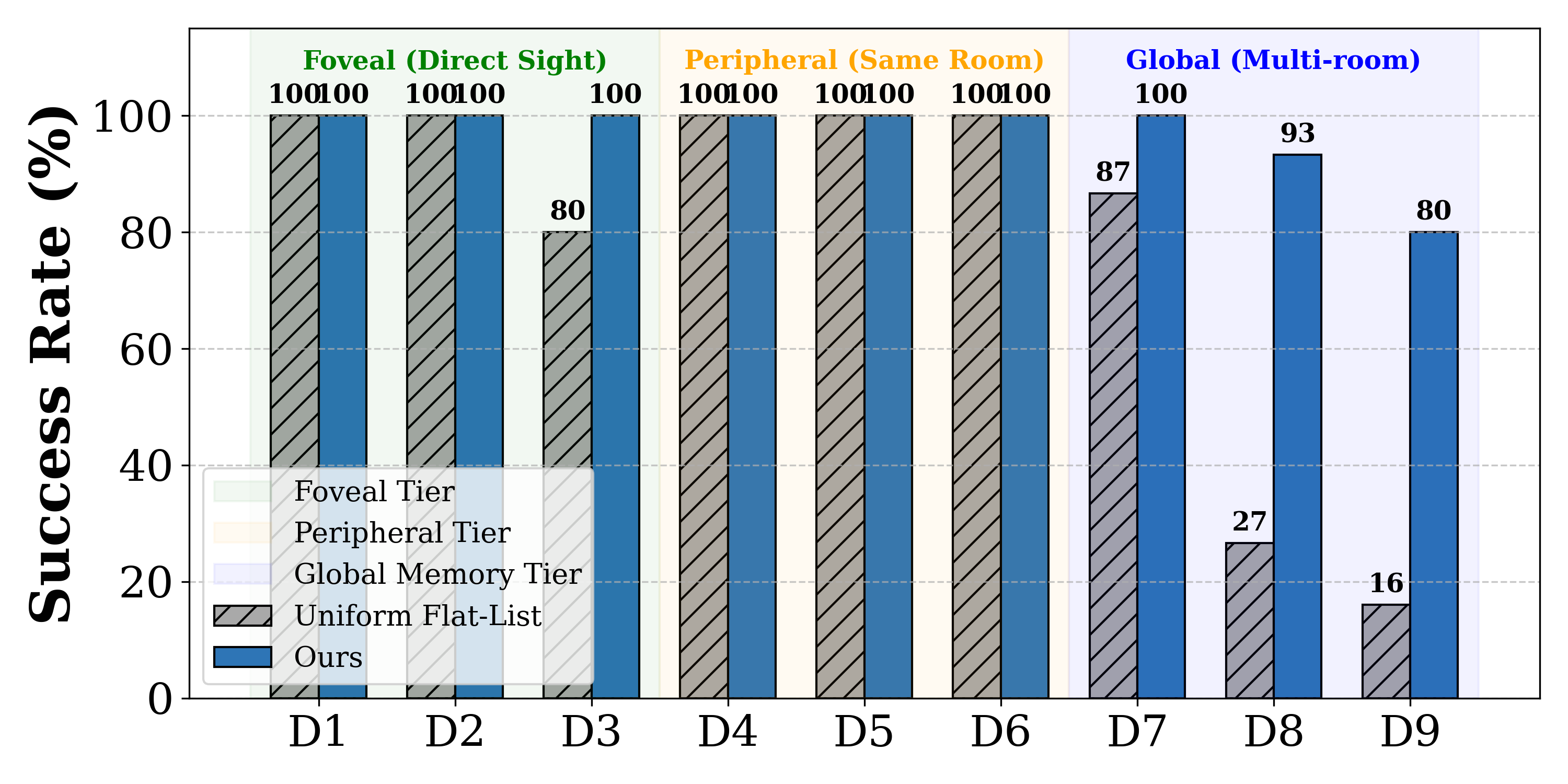}
    \caption{Navigation Success Rate (\%) comparison between Uniform Flat-List and our method across nine datasets ($D_1$-$D_9$). Background colors indicate the three spatial attention tiers.}
    \label{fig:nav_success_comparison}
\end{figure}

The multi-resolution mechanism improves decision accuracy alongside efficiency. 

Fig.~\ref{fig:nav_success_comparison} shows navigation success rates (SR) across nine datasets. In the foveal tier (D1-D3), our method achieves 100\% SR, whereas the baseline drops to 80\% in D3, suggesting that hierarchical anchoring reduces ambiguity even in direct sight. Both methods reach 100\% SR in the peripheral tier (D4-D6). The global memory tier (D7-D9) exhibits the largest performance gap. The flat-list baseline degrades to 16 percent SR in D9 due to the complexity of disambiguating long coordinate lists. By summarizing distant rooms into topological nodes, our method maintains an 80\% SR in D9, validating hierarchical abstraction as a reliable cognitive heuristic for long-horizon navigation.

\begin{figure}[t]
    \centering
    \includegraphics[width=\columnwidth]{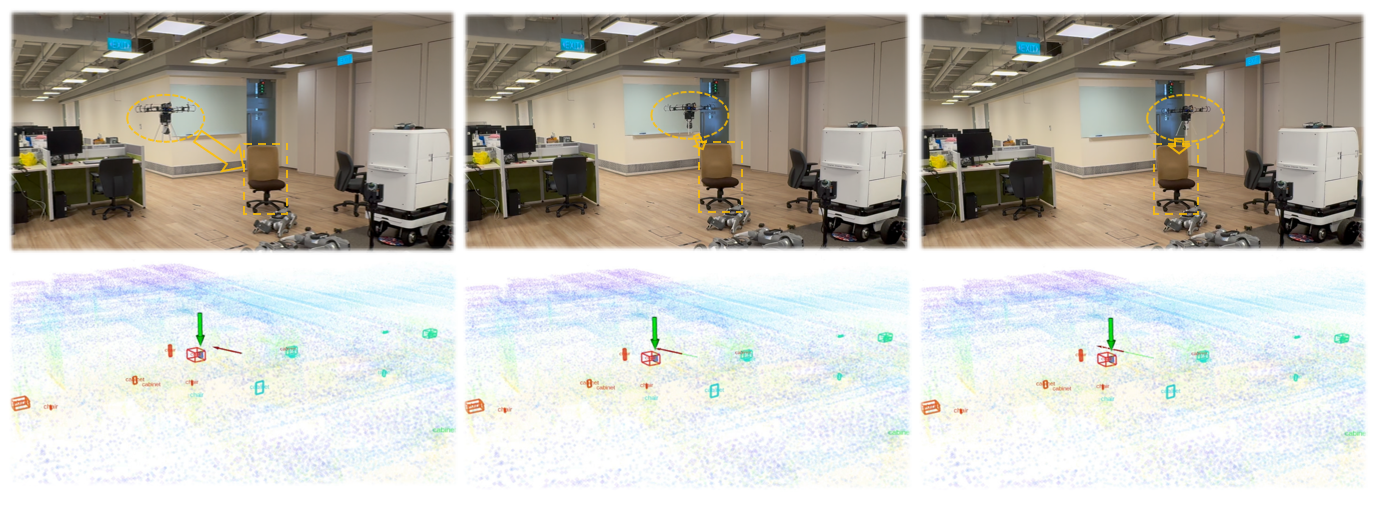}
    \caption{Qualitative zero-shot navigation results for the command ``go to the nearest chair''. The top row shows real-world execution, and the bottom row shows the synchronized 3D scene graph with target localization and the updated trajectory.}
    \label{fig:nav_execution}
\vspace{-0.2cm}
\end{figure}

\subsubsection{Impact on Inference Latency and Operational Scalability}

Multi-resolution prompting minimizes end-to-end latency by reducing input context length. Offloading spatial sorting to a Python-based geometry engine, which computes in under 10ms with zero token usage, significantly alleviates the LLM workload. This optimization is essential as inference time scales sensitively with prompt size.

Empirical tests in high-density scenarios demonstrate a reduction in average response time from 12.4s to 3.8s per decision compared to the flat-prompt baseline. This 3x speedup is critical for quadruped deployment, where excessive latency causes odometry drift and safety hazards. Faster reasoning maintains the tight perception-action loop required for real-time operation.

The architecture ensures operational scalability for long-horizon tasks by maintaining logarithmic prompt growth. While flat representations eventually hit context window limits, our DSG-based summarization decouples focal details from global memory. This approach sustains consistent performance in complex multi-room environments, offering a robust framework for scalable embodied AI.

\subsection{Real-world Aerial and Ground Platform Deployment and Testing}

OmniVLN validation uses FAST-LIO \cite{xu2022fast} for localization across aerial \cite{li2026aeos} and ground robots. Onboard modules manage perception and DSG construction, while LLM reasoning is handled offboard via WiFi. Hierarchical prompts, including DSG summaries and 3D octant observations, are transmitted to the remote model, with returned commands parsed into platform-specific setpoints.

The aerial platform integrates the PX4 autopilot \cite{meier2015px4}, and MINCO \cite{wang2022geometrically} trajectory optimizer. The quadruped uses the Unitree SDK and FAR local planner \cite{yang2022far} for collision-aware motion. Both systems share a unified reasoning pipeline, with platform-specific modifications limited to low-level control and local planning modules.

The architecture demonstrates high modularity by decoupling high-level DSG reasoning from low-level execution. This design enables OmniVLN to support heterogeneous robots with minimal modification to the reasoning stack. Such versatility confirms the framework's adaptability for zero-shot navigation across diverse hardware configurations in complex indoor scenes.

The robustness of this hierarchical reasoning process is qualitatively validated in Fig.~\ref{fig:nav_execution}, which demonstrates the seamless synchronization between real-world execution of the command ``go to the nearest chair'' and its corresponding localization and path generation within our 3D scene graph.

\subsection{Influence of Rotating LiDAR Scanning on Semantic Mapping Completeness}
\begin{figure}[t]
    \centering
    \subfloat[Rotating LiDAR system.\label{fig:map_rotating}]{\includegraphics[width=0.42\linewidth]{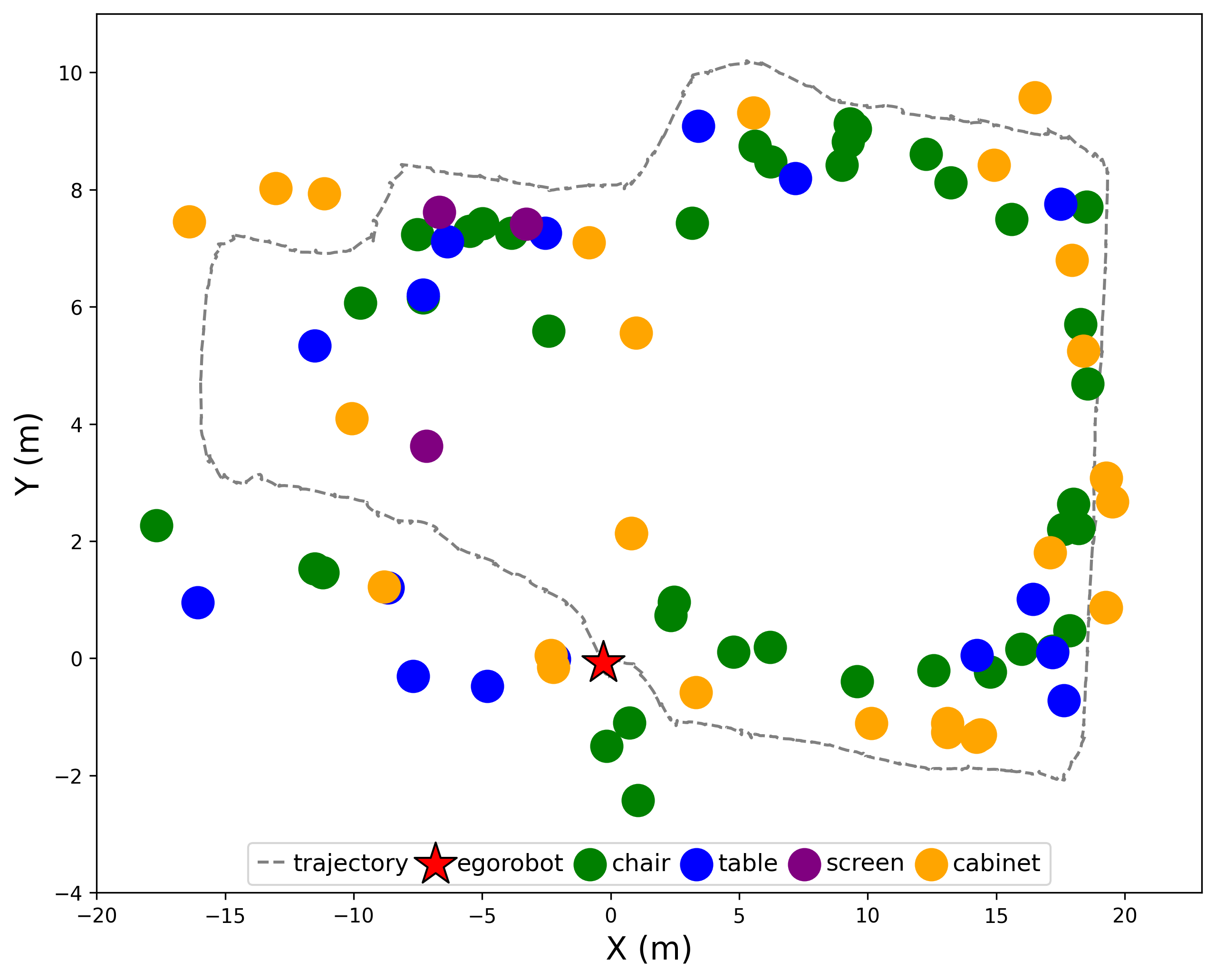}}
    \hfill
    \subfloat[Fixed LiDAR system.\label{fig:map_fixed}]{\includegraphics[width=0.42\linewidth]{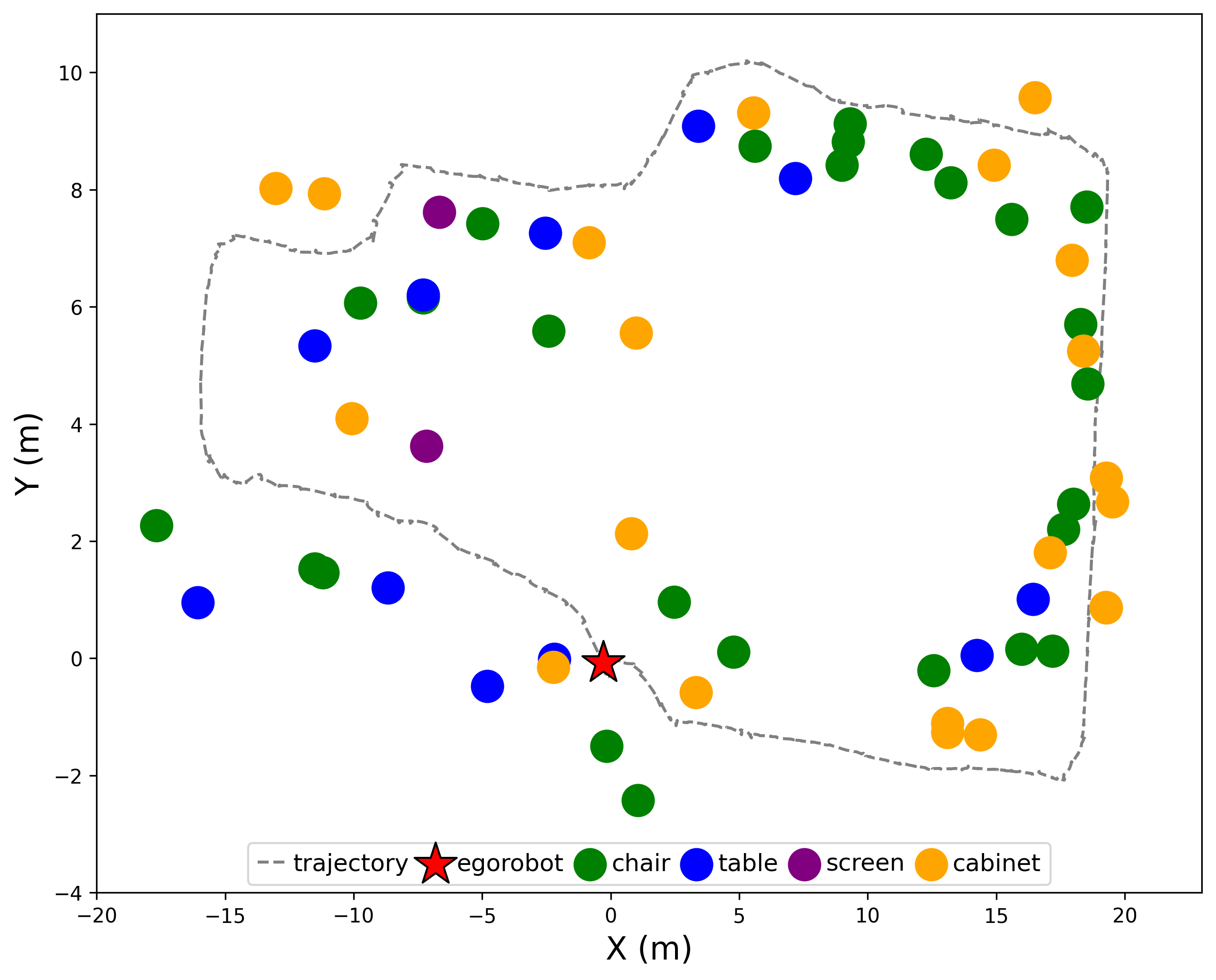}}
    \caption{Comparison of semantic mapping completeness under rotating and fixed LiDAR scanning configurations.}
    \label{fig:map_comparison}
\end{figure}

We further evaluate how LiDAR scanning patterns affect semantic map completeness and downstream scene graph construction. Under the same robot trajectory, environment, and semantic mapping pipeline, the rotating LiDAR configuration provides substantially more complete omnidirectional observations than the fixed LiDAR setting. Specifically, the rotating system instantiates 85 object nodes in the resulting semantic map, whereas the fixed LiDAR configuration identifies only 61 object nodes along the same path. As shown in Fig.~\ref{fig:map_comparison}, rotating LiDAR significantly improves coverage in side-facing and partially occluded regions, leading to denser semantic observations and a more structurally complete scene graph. These results confirm that omnidirectional active scanning alleviates perception lag and reduces topological fragmentation caused by limited sensor field of view.

\section{Conclusion}

This paper introduces OmniVLN, a zero-shot visual-language navigation framework merging omnidirectional perception with LLM-based spatial reasoning. Fusing rotating LiDAR and panoramic vision ensures spatio-temporal consistency while eliminating perception lag and topological fragmentation inherent in narrow-FOV systems. The proposed agent-centric 3D octant model discretizes surroundings into a volumetric 8-neighborhood using multi-resolution spatial attention, substantially reducing LLM token consumption without compromising navigation accuracy.

Future research will extend OmniVLN to dynamic environments by incorporating real-time pedestrian tracking and prediction into the scene graph hierarchy. Furthermore, we aim to investigate collaborative air-ground multi-agent systems where shared representations facilitate decentralized exploration and coordinated task allocation.

\bibliographystyle{IEEEtran}
\bibliography{refs} 

@article{xu2022fast,
  title={Fast-lio2: Fast direct lidar-inertial odometry},
  author={Xu, Wei and Cai, Yixi and He, Dongjiao and Lin, Jiarong and Zhang, Fu},
  journal={IEEE Transactions on Robotics},
  volume={38},
  number={4},
  pages={2053--2073},
  year={2022},
  publisher={IEEE}
}

@inproceedings{chaplot2020object,
  title={Object goal navigation using goal-oriented semantic exploration},
  author={Chaplot, Devendra Singh and Gandhi, Dhiraj Prakashchand and Gupta, Abhinav and Salakhutdinov, Russ R},
  booktitle={Advances in Neural Information Processing Systems (NeurIPS)},
  volume={33},
  pages={4247--4258},
  year={2020}
}

@inproceedings{mi2021object,
  title={Object goal navigation with end-to-end deep reinforcement learning},
  author={Mi, Haokun and Chen, Jing and others},
  booktitle={2021 IEEE/RSJ International Conference on Intelligent Robots and Systems (IROS)},
  pages={1234--1241},
  year={2021},
  organization={IEEE}
}

@article{wang2023pano,
  title={Pano-DRL: Panoramic Vision for Semantic Navigation in Cluttered Environments},
  author={Wang, X. and others},
  journal={IEEE Robotics and Automation Letters},
  year={2023}
}

@inproceedings{zhang2014loam, title={{LOAM}: Lidar Odometry and Mapping in Real-time}, author={Zhang, Ji and Singh, Sanjiv}, booktitle={Robotics: Science and Systems (RSS)}, year={2014}, address={Berkeley, CA} }

@article{geyer2001ijcv,
  author  = {Geyer, Christopher and Daniilidis, Kostas},
  title   = {Catadioptric Projective Geometry},
  journal = {International Journal of Computer Vision},
  volume  = {45},
  number  = {3},
  pages   = {223--243},
  year    = {2001},
  doi     = {10.1023/A:1013610201135}
}

@inproceedings{scaramuzza2006iros,
  author    = {Scaramuzza, Davide and Martinelli, Agostino and Siegwart, Roland},
  title     = {A Toolbox for Easily Calibrating Omnidirectional Cameras},
  booktitle = {IEEE/RSJ International Conference on Intelligent Robots and Systems (IROS)},
  year      = {2006}
}

@inproceedings{geiger2012cvpr,
  author    = {Geiger, Andreas and Lenz, Philip and Urtasun, Raquel},
  title     = {Are we ready for Autonomous Driving? The KITTI Vision Benchmark Suite},
  booktitle = {IEEE Conference on Computer Vision and Pattern Recognition (CVPR)},
  pages     = {3354--3361},
  year      = {2012}
}

@InProceedings{Singh_2023_ICCV,
    author    = {Singh, Apoorv},
    title     = {Surround-View Vision-Based 3D Detection for Autonomous Driving: A Survey},
    booktitle = {Proceedings of the IEEE/CVF International Conference on Computer Vision (ICCV) Workshops},
    month     = {October},
    year      = {2023},
    pages     = {3243-3252}
}

@article{li2026aeos,
  title={Aeos: Active environment-aware optimal scanning control for uav lidar-inertial odometry in complex scenes},
  author={Li, Jianping and Xu, Xinhang and Liu, Zhongyuan and Yuan, Shenghai and Cao, Muqing and Xie, Lihua},
  journal={ISPRS Journal of Photogrammetry and Remote Sensing},
  volume={232},
  pages={476--491},
  year={2026},
  publisher={Elsevier}
}

@inproceedings{armeni20193d,
  title={3d scene graph: A structure for unified semantics, 3d space, and camera},
  author={Armeni, Iro and He, Zhi-Yang and Gwak, JunYoung and Zamir, Amir R and Fischer, Martin and Malik, Jitendra and Savavese, Silvio},
  booktitle={Proceedings of the IEEE/CVF International Conference on Computer Vision (ICCV)},
  pages={5664--5673},
  year={2019}
}

@article{rosinol2021kimera,
  title={Kimera: from SLAM to spatial perception with 3D dynamic scene graphs},
  author={Rosinol, Antoni and Violette, Andrew and Abate, Marcus and Chang, Yun and Shi, Jingnan and Gupta, Ali and Carlone, Luca},
  journal={The International Journal of Robotics Research},
  volume={40},
  number={12-14},
  pages={1510--1546},
  year={2021},
  publisher={SAGE Publications}
}

@inproceedings{zhou2023esc,
  title={Esc: Exploration with soft commonsense constraints for zero-shot object navigation},
  author={Zhou, Kaiwen and Zheng, Kaizhi and Pryor, Connor and Shen, Yilin and Jin, Hongxia and Getoor, Lise and Wang, Xin Eric},
  booktitle={International Conference on Machine Learning (ICML)},
  pages={42829--42842},
  year={2023},
  organization={PMLR}
}

@article{ahn2022saycan,
  title={Do as i can, not as i say: Grounding language models in robotic affordances},
  author={Ahn, Michael and Brohan, Anthony and Brown, Noah and Chebotar, Yevgen and Cortes, Omar and David, Byron and Finn, Chelsea and Fu, Chuyuan and Gopalakrishnan, Keerthana and Hausman, Karol and others},
  journal={arXiv preprint arXiv:2204.01691},
  year={2022}
}

@article{dorbala2024can,
  title={Can LLMs Plan and Execute ObjectNav? Redefining the Zero-Shot Object Navigation Pipeline},
  author={Dorbala, Vishnu Sashank and others},
  journal={arXiv preprint arXiv:2401.02115},
  year={2024}
}

@article{liu2024llava,
  title={Visual instruction tuning},
  author={Liu, Haotian and Li, Chunyuan and Wu, Qingyang and Lee, Yong Jae},
  journal={Advances in Neural Information Processing Systems (NeurIPS)},
  volume={36},
  year={2024}
}

@article{qwen25,
  title={Qwen2.5 Technical Report},
  author={An, Yang and Cheng, Baite and Chen, Bo and Chen, Jianghao and Chen, Zheng and others},
  journal={arXiv preprint arXiv:2412.15115},
  year={2024}
}

@article{zantout2025sort3d,
  title={SORT3D: Spatial Object-centric Reasoning Toolbox for Zero-Shot 3D Grounding Using Large Language Models},
  author={Zantout, Nader and Zhang, Haochen and Kachana, Pujith and Qiu, Jinkai and Chen, Guofei and Zhang, Ji and Wang, Wenshan},
  journal={arXiv preprint arXiv:2504.18684},
  year={2025}
}

@article{hughes2023foundations,
  title={Foundations of Spatial Perception for Robotics: Hierarchical Representations and Real-time Systems},
  author={Hughes, Nathan and Chang, Yun and Hu, Siyi and Talak, Rajat and Abdulhai, Rumaisa and Strader, Jared and Carlone, Luca},
  journal={The International Journal of Robotics Research},
  year={2023},
  publisher={SAGE Publications}
}

@article{xu2025s2mllm,
  title={S$^2$-MLLM: Boosting Spatial Reasoning Capability of MLLMs for 3D Visual Grounding with Structural Guidance},
  author={Xu, Beining and Zhu, Siting and Jin, Zhao and Li, Junxian and Wang, Hesheng},
  journal={arXiv preprint arXiv:2512.01223},
  year={2025}
}

@article{mi2025lasp,
  title={Language-to-Space Programming for Training-Free 3D Visual Grounding},
  author={Mi, Boyu and Wang, Hanqing and Wang, Tai and Chen, Yilun and Pang, Jiangmiao},
  journal={Proceedings of the 2025 Conference on Empirical Methods in Natural Language Processing (EMNLP)},
  pages={3844--3864},
  year={2025}
}

@inproceedings{yin2024sgnav,
  title={SG-Nav: Online 3D Scene Graph Prompting for LLM-based Zero-shot Object Navigation},
  author={Yin, Hang and Xu, Xiuwei and Wu, Zhenyu and Jie, Zhou and Lu, Jiwen},
  booktitle={Advances in Neural Information Processing Systems (NeurIPS)},
  year={2024}
}

@inproceedings{dai2017scannet,
  title={Scannet: Richly-annotated 3d reconstructions of indoor scenes},
  author={Dai, Angela and Chang, Angel X and Savva, Manolis and Halber, Maciej and Funkhouser, Thomas and Nie{\ss}ner, Matthias},
  booktitle={Proceedings of the IEEE conference on computer vision and pattern recognition},
  pages={5828--5839},
  year={2017}
}

@inproceedings{chang2017matterport3d,
  title={Matterport3d: Learning from rgb-d data in indoor environments},
  author={Chang, Angel and Dai, Angela and Funkhouser, Thomas and Halber, Maciej and Niessner, Matthias and Savva, Manolis and Song, Shuran and Zeng, Andy and Zhang, Yinda},
  booktitle={International Conference on 3D Vision (3DV)},
  year={2017}
}

@inproceedings{deitke2020robothor,
  title={Robothor: An open simulation-to-real embodied ai platform},
  author={Deitke, Matt and Han, Winson and Herrasti, Alvaro and Kembhavi, Aniruddha and others},
  booktitle={Proceedings of the IEEE/CVF conference on computer vision and pattern recognition},
  pages={3164--3174},
  year={2020}
}

@article{ren2024grounded,
  title={Grounded SAM: Assembling Open-World Models for Diverse Visual Tasks},
  author={Ren, Tianhe and Shilong, Liu and others},
  journal={arXiv preprint arXiv:2401.14159},
  year={2024}
}

@article{zhang2026spatialnav,
  title={SpatialNav: Leveraging Spatial Scene Graphs for Zero-Shot Vision-and-Language Navigation},
  author={Zhang, Jiwen and Li, Zejun and Wang, Siyuan and Shi, Xiangyu and Wei, Zhongyu and Wu, Qi},
  journal={arXiv preprint arXiv:2601.06806},
  year={2026}
}

@inproceedings{savva2019habitat,
  title={Habitat: A Platform for Embodied {AI} Research},
  author={Savva, Manolis and Kadian, Abhishek and Maksymets, Oleksandr and Zhao, Yili and Wijmans, Erik and Jain, Bhavana and Straub, Julian and Liu, Jia and Koltun, Vladlen and Malik, Jitendra and others},
  booktitle={Proceedings of the IEEE/CVF International Conference on Computer Vision (ICCV)},
  pages={9339--9347},
  year={2019}
}

@inproceedings{xia2018gibson,
  title={Gibson {Env}: Real-World Perception for Embodied Agents},
  author={Xia, Fei and Zamir, Amir R and He, Zhiyang and Sax, Alexander and Malik, Jitendra and Savarese, Silvio},
  booktitle={Proceedings of the IEEE Conference on Computer Vision and Pattern Recognition (CVPR)},
  pages={9068--9079},
  year={2018}
}

@inproceedings{meier2015px4,
  title={PX4: A node-based multithreaded open source robotics framework for deeply embedded platforms},
  author={Meier, Lorenz and Honegger, Dominik and Pollefeys, Marc},
  booktitle={2015 IEEE international conference on robotics and automation (ICRA)},
  pages={6235--6240},
  year={2015},
  organization={IEEE}
}

@article{wang2022geometrically,
  title={Geometrically constrained trajectory optimization for multicopters},
  author={Wang, Zhepei and Zhou, Xin and Xu, Chao and Gao, Fei},
  journal={IEEE Transactions on Robotics},
  volume={38},
  number={5},
  pages={3259--3278},
  year={2022},
  publisher={IEEE}
}

@inproceedings{yang2022far,
  title={Far planner: Fast, attemptable route planner using dynamic visibility update},
  author={Yang, Fan and Cao, Chao and Zhu, Hongbiao and Oh, Jean and Zhang, Ji},
  booktitle={2022 ieee/rsj international conference on intelligent robots and systems (iros)},
  pages={9--16},
  year={2022},
  organization={IEEE}
}

\end{document}